\documentclass[letterpaper, 10 pt, conference]{ieeeconf}  

\IEEEoverridecommandlockouts                              

\usepackage{graphicx}
\usepackage{wrapfig}
\usepackage[x11names]{xcolor}
\usepackage{bbm}
\usepackage{titletoc}

\usepackage{amsmath,amsfonts,bm}

\def\eqref#1{equation~\ref{#1}}

\def\1{\bm{1}}

\DeclareMathAlphabet{\mathsfit}{\encodingdefault}{\sfdefault}{m}{sl}
\SetMathAlphabet{\mathsfit}{bold}{\encodingdefault}{\sfdefault}{bx}{n}

\usepackage{hyperref}
\usepackage{url}

\usepackage{amsmath}
\usepackage{cleveref}
\usepackage{multirow}
\usepackage{booktabs}
\usepackage{color}
\usepackage{colortbl}
\usepackage{algorithm}
\usepackage{algorithmicx}
\usepackage{algpseudocode}
\usepackage{subcaption}
\usepackage{sidecap}
\usepackage{soul}
\usepackage{graphicx}
\usepackage{adjustbox}
\usepackage{booktabs}
\usepackage{dsfont}

\definecolor{lightblue}{rgb}{0.678, 0.847, 0.902}
\definecolor{lightgreen}{rgb}{01.0, 0.8, 0.6}

\title{IGDrivSim: A Benchmark for the Imitation Gap in Autonomous Driving}

\author{%
    \textbf{Clémence Grislain}\\
    Sorbonne University$^*$ \thanks{$^*$Work done while interning at University of Oxford}\\
    \texttt{clemgris@gmail.com}
    \and
    \textbf{Risto Vuorio} \\
    University of Oxford\\
    \and
    \textbf{Cong Lu}\\
    University of British \\ Columbia,
    Vector Institute \\
    \and
    \textbf{Shimon Whiteson}\\
    University of Oxford\\
}

\begin{document}

\maketitle

\begin{abstract}
Developing autonomous vehicles that can navigate complex environments with human-level safety and efficiency is a central goal in self-driving research. A common approach to achieving this is imitation learning, where agents are trained to mimic human expert demonstrations collected from real-world driving scenarios. However, discrepancies between human perception and the self-driving car's sensors can introduce an \textit{imitation gap}, leading to imitation learning failures. In this work, we introduce \textbf{IGDrivSim}, a benchmark built on top of the Waymax simulator, designed to investigate the effects of the imitation gap in learning autonomous driving policy from human expert demonstrations. Our experiments show that this perception gap between human experts and self-driving agents can hinder the learning of safe and effective driving behaviors.
We further show that combining imitation with reinforcement learning, using a simple penalty reward for prohibited behaviors, effectively mitigates these failures. Our code is open-sourced at: \url{https://github.com/clemgris/IGDrivSim.git}.
\end{abstract}

\section{Introduction}

Research in autonomous driving aims to develop vehicles capable of navigating complex environments with the safety and efficiency of human drivers~\cite{levinson2011towards, yurtsever2020survey}.
Learning such autonomous driving policies has been a focus of the research community due to its potential to cut costs, ease labor constraints, improve safety, and enhance mobility \cite{maurer2016autonomous}. However, achieving this level of performance remains a critical challenge in the field. In fact, while reinforcement learning (RL,~\cite{sutton2018reinforcement}) has been a natural solution for control in various domains \cite{mnih2013playing}, self-driving agents often struggle to learn desired behaviors using RL due to the high computational cost and, more critically, the difficulty of designing an effective reward function. In these scenarios, imitation learning from human driver demonstrations (IL,~\cite{schaal1999is,hussein2017imitation}) is a popular alternative for training policies. IL allows self-driving agents to learn directly from human expert behaviors, bypassing the need for an explicit reward function. In the literature, expert datasets often consist of traffic scenarios where multiple road users are tracked in 3D space using sensors from a human-driven car~\cite{dosovitskiy2017carla, amini2021vista, ettinger2021large}. These data are relatively simple to record in real-world environments, and training typically involves simulators that replicate these recorded driving scenarios, with the task being to predict the trajectories of each agent throughout the scene. 

However, a major challenge of IL arises when the imitator and the expert have different \textit{observability} of the environment, which can occur, for instance, when they rely on different sensors. This difference can lead to an \textit{imitation gap}~\cite{swamy2021moments, weihs2021bridging, walsman2023impossibly, vuorio2024bayesian} and potentially cause IL algorithms to fail. Imagine training an autonomous vehicle to navigate in a foggy area by imitating data from another vehicle that was recorded in clear weather. The expert vehicle, operating in clear conditions, can see far ahead and navigate smoothly, anticipating obstacles and changes in the environment well in advance. However, when the imitator behaves in an area with dense fog, it cannot see as far ahead and must rely solely on its limited visibility to navigate. If the imitator simply copies the expert's actions, driving at the same speed and making the same turns, it will likely fail because it does not have the same information about the road ahead. Instead, a better policy for the imitator would involve slowing down and adopting a less smooth but more reactive behavior to avoid obstacles. This cautious and adaptive behavior allows the imitator to compensate for its reduced visibility. However, this behavior is never demonstrated by the expert vehicle and therefore is not learned by the imitator.
\begin{figure}
\centering
\includegraphics[width=\columnwidth]{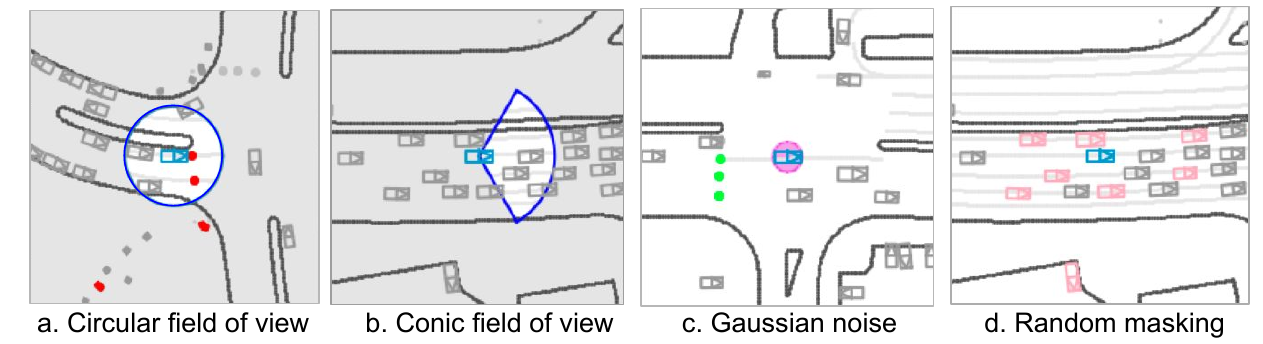}
\caption{
\small \textbf{IGDrivSim} scenarios illustrating partial observability: the blue vehicle navigates among road lines, streetlights, and other vehicles. From left to right: (a) Circular field of view (blue), (b) Conic field of view (blue), (c) Noisy self-position (magenta), (d) Noisy vehicle detection (occluded vehicles in pink).}
\vspace{-6mm}
\label{fig:benchmark}
\end{figure}

In this example, the difference in observability between the expert and the imitator would cause traditional IL algorithms to fail. Critically, the common practice of training autonomous driving policies using simulators based on human driver trajectories naturally introduces a similar difference in observability between the human expert and the self-driving car (SDC), which could likewise result in the failure of IL. In fact, when performing IL, the SDC has access to all the information recorded during data collection, typically involving sensors like LiDAR and cameras. Yet, these sensors differ in their capabilities from human perception, on which the human driver relies to generate expert demonstrations. For example, a human driver might react based on audio cues like honks that the SDC might not detect if it lacks audio sensors. In contrast, human drivers might not see certain information due to visual blind spots that the SDC would detect with its broader field of view. These differences in the perception of the driving environment between the human expert and the SDC imitator create an imitation gap. Imitation learning from human expert demonstrations may lead the SDC imitator to not learn the specific behaviors required for its particular sensory capabilities (i.e., the sensors used to collect scene information). In the best case, the learned policy is safe but suboptimal under the SDC's observability; in the worst case, it learns unsafe behaviors. Although the imitation gap has been studied in previous work~\cite{weihs2021bridging, walsman2023impossibly, vuorio2024bayesian}, to our knowledge, there exists no dedicated benchmark for evaluating how this difference in observability affects the learning of driving policies from human demonstrations. 

Therefore, we introduce \textbf{IGDrivSim} (\textbf{I}mitation \textbf{G}ap in \textbf{Driv}ing \textbf{Sim}ulation), a benchmark specifically designed to explore the impact of the imitation gap on learning driving policies from human demonstrations. IGDrivSim introduces constraints (see \Cref{fig:benchmark}) on the observability of the Waymax driving simulator~\cite{gulino2023waymax}, which is based on the WOMD dataset~\cite{ettinger2021large} of real-world human driving data. These constraints amplify the difference between the human expert's and the simulated self-driving car's observability, effectively widening the imitation gap. Our experiments reveal that when trained with IL, especially behavioral cloning, a self-driving agent —whose observations of the driving scene inherently differ from human perception— struggles to learn good behaviors from human expert demonstrations alone and that this failure can be explained by the presence of an imitation gap. 
Furthermore, we demonstrate that incorporating knowledge about prohibited behaviors can help address this issue.
In fact, we find that by combining IL with RL based on a simple penalty reward, we can mitigate these failures. While Lu et al.~\cite{lu2023imitation} demonstrated that reward signals can enhance IL, the underlying reasons were not fully explored.
In this work, we present a novel setup and results that show how an imitation gap can explain the need for supplemental methods like RL in training self-driving policies with IL. By introducing IGDrivSim as a benchmark, we aim to provide a tool for evaluating and addressing the imitation gap in autonomous driving, facilitating both the assessment and improvement of IL from human driving~demonstrations.

In summary, our contributions are threefold:
\begin{enumerate}
    \item We release an open-source benchmark, called IGDrivSim, for studying the imitation gap in a high-dimensional, real-world complex task where IL is the current state-of-the-art method for training policies.
    \item We highlight the importance of addressing the imitation gap when training self-driving car policies on human driver data, revealing key limitations of demonstration-based IL.
    \item We release the first (to our knowledge) open-source motion prediction baselines trained on the Jax-based simulator Waymax.
\end{enumerate}

\section{Related Work}
\label{sec:related_work}
A variety of methods have been proposed recently to solve the imitation gap, but a lack of standardized benchmarks makes comparison among the algorithms harder. We provide a principled imitation gap benchmark with real-world motivation from learning policies for self-driving cars.

A popular category of algorithms for solving the imitation gap combines BC and RL objectives using different weighting schemes. Weihs et al.~\cite{weihs2021bridging} propose an algorithm that dynamically weights the objectives during training, while Nguyen et al.~\cite{nguyen2022leveraging} start from an offline RL algorithm accelerated by the BC objective. We align with this line of research and consider a baseline that combines BC and RL with fixed weights to address the imitation gap. Another solution idea is to use DAgger-style~\cite{Ross2011-kt} online supervision to accelerate learning in an imitation gap setting~\cite{walsman2023impossibly}. 

Vuorio et al.~\cite{vuorio2024bayesian} show that prior information about the desired behaviors is always required to solve the imitation gap and propose a Bayesian solution to the problem.
The failure of BC in problems with an imitation gap can be seen as a form of causal confounding~\cite{Ortega2021-hh}.
When the imitator can solve the problem without deviating from the expert behavior, Ortega et al. \cite{Ortega2021-hh} and Swamy et al.~\cite{swamy2022sequence} show that DAgger-style methods can work well in this setting, while Vuorio et al.~\cite{vuorio2024deconfounding} use a separate inference model. In our work, we address scenarios where imitators must deviate from the expert to achieve optimal behavior and propose to use additional information from prohibited behaviors to mitigate the imitation gap problem.


Learning driving policies is a popular field of research with multiple datasets and benchmarks targeting it.
Waymo Open Dataset (WOMD)~\cite{ettinger2021large} contains many driving scenarios with log trajectories of human drivers.
The Waymax~\cite{gulino2023waymax} simulator we build upon runs on the scenarios from the WOMD dataset.
Argoverse2~\cite{wilson2023argoverse} provides an alternative to WOMD, while NuPlan~\cite{caesar2021nuplan} is one for Waymax.
We chose Waymax due to its fast accelerator-compatible implementation and the large quantity of high-quality data included in WOMD. A closely related benchmark to ours is \cite{vinitsky2023nocturne}, which builds on WOMD and features partial observability.
However, the effect of partial observability in Nocturne is unknown as it is not ablated. Perhaps most closely related to ours is ``Learning by Cheating'' by Chen et al.~\cite{chen2019lbc}, where they consider a vision-based driving setting with an imitation gap problem based on the CARLA driving simulator~\cite{dosovitskiy2017carla}.

BC is a popular method for learning driving policies.
Seff et al.~\cite{seff2023motionlm} propose to train a transformer policy on WOMD using BC.
Lu et al.~\cite{lu2023imitation} demonstrate that  BC is not always enough and show that combining BC with RL can improve metrics on the WOMD data.
The current leader on the Waymo Open Sim Agents Challenge~\cite{montali2023waymo}, which also uses WOMD data is also a transformer policy trained with BC~\cite{wu2024smart}.

\section{Background}
\label{sec:background}

\subsection{Preliminaries}
\label{subsec:preliminaries}

\noindent \paragraph{\textbf{Reinforcement Learning}} We define the environment as a Partially Observable Markov Decision Process (POMDP)~\cite{spaan2012partially}, characterized by the tuple $\mathcal{M} := \left<\mathcal{S}, \mathcal{A}, p(s_{t+1}|s_t, a_t), r(s_t, a_t), \Omega, O(s_t), \gamma \right>$. Similar to the classic MDP definition, $\mathcal{S}$ and $\mathcal{A}$ denote the state and action spaces respectively, $p(s_{t+1}|s_t, a_t)$ represents the transition dynamics, and $r(s_t, a_t)$ is the reward function. However, in a POMDP, the agent observes the environment through an observation function $O: \mathcal{S}\rightarrow \mathcal{P}(\Omega)$, which maps states to probability distributions over the observation space $\Omega$.
In RL, the objective is to optimize a policy $\pi(a | z)$ with $z\sim O(s)$ that maximizes the expected return $\mathbb{E}_{\pi, p}\left[\sum_{t=0}^\infty \gamma^t r(s_t, a_t)\right]$.

\noindent \paragraph{\textbf{Behavioral Cloning}}
In IL, we assume access to a set of expert demonstrations $\mathcal{D}_\textrm{expert}:= \{\tau_i\}_{i=1}^{N_\textrm{Expert}}$ of state-action trajectories $\tau_i\:= \{s_0,a_0,s_1,a_1,\dots \}$ from which an imitator learns a behavior. Behavioral cloning (BC,~\cite{pomerleau1998alvinn}) is an IL method where the objective is to learn a policy $\pi(a|z)$ with $z \sim O(s)$ that maximizes the log-likelihood of the expert actions with respect to the action distribution predicted by the imitator along the expert's trajectory $\sum_{\tau_i \in \mathcal{D}_\text{expert}} \sum_{(a_j,s_j)\in\tau_i}\log \pi(a_j|z_j)$ with $z_j \sim O(s_j)$. 

\noindent \paragraph{\textbf{Imitation Gap Problem}} 
An imitation gap arises when the demonstrator and the imitator have different observation functions, i.e., $O_\text{expert} \neq O_\text{imitator}$. A sufficient condition for the imitation gap to cause IL failure (i.e., leading to a suboptimal policy as described in \cite{swamy2021moments}) is when the expert's policy, which generates the expert demonstrations, becomes suboptimal in the imitator's POMDP. In other words, the expert's policy, which is assumed to be Bayes-optimal \cite{ross2007bayes} in the expert's POMDP, is no longer Bayes-optimal in the imitator's POMDP, due to the imitation gap.

As a result, the imitator may need to deviate from the expert to learn an optimal policy in its own POMDP. However, standard demonstration-based IL algorithms do not permit such deviations. For instance, BC minimizes the discrepancy between the expert’s and the imitator’s state-action pairs, constraining the imitator to closely follow the expert's behavior. While this work uses BC to illustrate IL failures under the imitation gap, other demonstration-based IL algorithms face similar issues, as shown by \cite{weihs2021bridging} for Generative Adversarial Imitation Learning (GAIL,~\cite{ho2016generative}) in simple environments. To address the imitation gap and enable the imitator to learn an optimal policy, additional guidance is required. This can be provided through an explicit Bayesian prior \cite{vuorio2024bayesian}, which leverages prior knowledge, or through the environment's reward signal \cite{weihs2021bridging}, which can help align the imitator’s learning process with optimal behavior.

\subsection{Waymax Simulator}
\label{subsec:waymax}

Waymax is a Jax-based data-driven simulator built on the WOMD dataset, which consists of trajectories of vehicles, road users, and features in diverse urban settings. WOMD contains over 570 hours of data from 1,750 km of roads across six U.S. cities. The data consists of 9.1 seconds of driving scenarios mined for extracting interactive behaviors. Each scenario has 91 timesteps, the first 10 for context and the remaining 81 for behavior prediction. The task is to predict future vehicle trajectories.

\subsection{Action Space}
The Waymax simulator introduces a \textit{delta} car control model, where the action space is continuous and defined by the position and yaw displacements $(\Delta x, \Delta y, \Delta \theta)$ of the agent between two consecutive states. When an agent in position and yaw $(x,y,\theta)$ takes action $(\Delta x, \Delta y, \Delta \theta)$ it transitions to $(x+\Delta x, y+\Delta y, \theta+\Delta \theta)$. Its speed is also updated to $v_x=\Delta x / \Delta t$ and $v_y =\Delta_y / \Delta t$. This action space is advantageous as the inverse kinematics used to derive the expert's actions from the logged trajectory, required for BC in particular, are straightforward.

\subsection{State Space}
The Waymax state space includes map data—polylines with road feature types—and scene object information. Objects are represented as bounding boxes and velocity vectors. The imitator learns from the ego car's trajectory, perceiving the environment from its local frame. Scene object data is limited to \((x,y,z,v,yaw)\), where \((x,y,z) \in \mathbb{R}^3\) represents position, \(v \in \mathbb{R}_+\) velocity, and \(yaw \in [0, 2\pi]\) heading.

\section{Mitigating the Imitation Gap}
\label{sec:method}

\subsection{Imitation Learning Failure under the Imitation Gap}
\label{subsec:minigrid}


To provide insight into why IL can fail in the presence and imitation gap, we examine a simple navigation problem where an agent has to reach a goal within a maze in the fewest possible timesteps. In this environment, we define two POMDPs for the expert and the imitator, differing in their observation functions. The imitator's observation function, defined by the agent's receptive field, masks the goal's position when it falls outside its field, whereas the expert always accesses the goal position.

In this setting, the expert demonstrates only goal-directed behaviors. However, for the imitator to solve the task, the optimal policy requires exploring the environment first to locate the goal. When trained on demonstrations generated by the expert with full observability, the imitator will never learn the necessary exploration behavior, resulting in a suboptimal policy under the imitator's POMDP. To characterize this failure, we approximate the Bayes-optimal policy for the imitator’s POMDP using PPO \cite{schulman2017proximal} and show that this policy cannot be achieved by training the imitator with BC on demonstrations from an expert with full observability.

\begin{figure}
\centering
\includegraphics[width=1.\columnwidth]{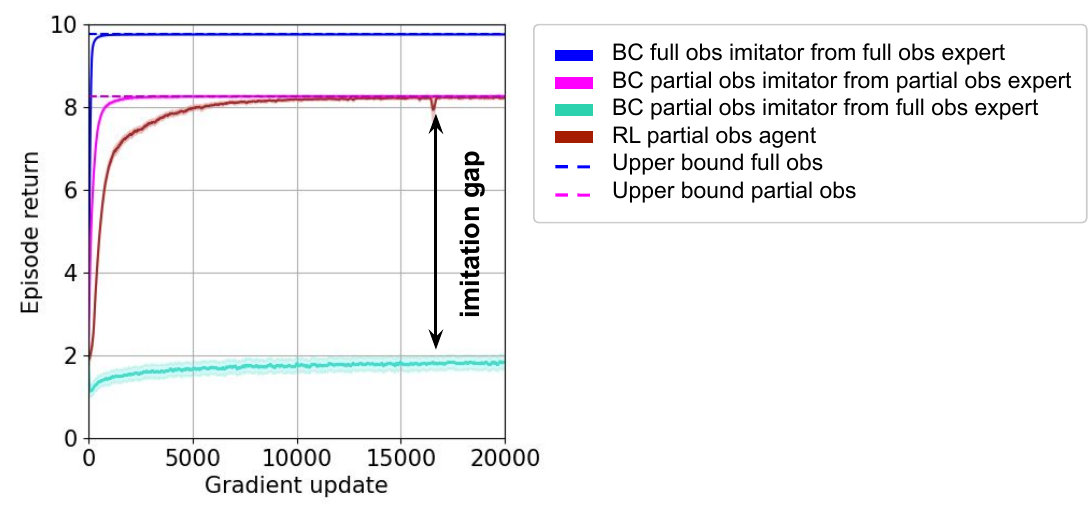}
\caption{
\small
\textbf{BC fails under the imitation gap.} Solid lines show episodic returns for agents trained with RL or BC under full or partial observability (receptive field size 3). Dashed lines indicate upper bound returns from converged RL agents approximating Bayes-optimal policies, and shading represents the mean and 95\% confidence interval across 10 seeds.
}
\vspace{-4mm}
\label{fig:maze_expe}
\end{figure}

\Cref{fig:maze_expe} presents the results of this experiment in a $13\times 13$ maze with a receptive field size of 3 for the imitator. The results indicate that the imitator with partial observability, trained on demonstrations from the expert with full observability (in turquoise), cannot learn an optimal policy w.r.t.\ its observability, and achieves suboptimal returns. On the other hand, when the expert and imitator share the same level of observability—either full or partial, respectively in blue and magenta--the imitator quickly learns an optimal policy with BC and achieves the maximum return. While in this environment the ground truth reward is known, which allows an optimal policy to be learned through RL (in brown), in many real-world scenarios, like autonomous driving, such reward fully defining the desired behavior is often inaccessible. Therefore, learning policy relies heavily on IL, hence the importance of addressing the imitation gap.

\subsection{Mitigating the Imitation Gap Problem with RL}
\label{subsec:BC_RL}

The failure of IL under mismatched expert and imitator observabilities occurs because the expert demonstrates behaviors that are optimal for the expert's POMDP but not for the imitator's POMDP. Consequently, for the imitator to learn behaviors optimal for its own partial observability, it must rely on signals outside the expert demonstration dataset. Although IL is typically employed in environments where defining a reward function for training a good policy is impractical, it is often possible to access a reward function that penalizes undesirable behaviors, such as safety metrics in autonomous driving.

While this reward may not fully characterize the desired behavior, and training solely based on it is impractical, it can still be used as an additional signal. This idea has been exploited by previous work~\cite{weihs2021bridging} as a solution for mitigating the imitation gap problem.
Following a similar approach to Lu et al.\cite{lu2023imitation}, we introduce a combined loss that leverages BC to learn generally useful behaviors from the expert and RL to learn behaviors specific to the imitator's observability. Here, the RL loss acts as a regularizer for the BC loss:
\begin{equation}
\label{eq:loss}
\mathcal{L}(\phi) = \omega_{BC} \mathcal{L}_{BC}(\phi, \mathcal{D}_\text{expert}) + \omega_{RL} \mathcal{L}_{RL}(\phi), \end{equation}
where $\phi$ represents the parameters of the imitator policy, and $(\omega_{BC}, \omega_{RL})$ are two hyperparameters. The BC loss is the negative log-likelihood of the expert's actions with respect to the action distribution predicted by the imitator policy, as defined in Section~\ref{subsec:preliminaries}. While we use BC loss, other standard IL algorithms and their respective loss functions can be used. The RL loss can also be derived from any RL algorithm; in this work, we use PPO~\cite{schulman2017proximal}.


\subsection{Highlighting the Imitation Gap in Waymax}
\label{subsec:obs_waymax}


To highlight the imitation gap problem, we introduce IGDrivSim, a benchmark that integrates imitators with partial observability into the Waymax simulator. IGDrivSim defines two categories of partial observability: limited field of view and noisy perception. In the limited field of view setting, observation functions obscure road and vehicle features outside a defined region, modeled as either a circular region with radius \(r \in \mathbb{R}_+\) or a conic region with opening angle \(\theta \in \mathbb{R}_+\) and radius \(r\) (see \Cref{fig:benchmark}). In the noisy perception setting, sensor noise alters the agent’s view of the environment. Specifically, Gaussian noise with zero mean and standard deviation \(\sigma\) is applied to the ego car’s global position, affecting all other vehicles and road points in the ego frame. Additionally, agents may experience random masking, where vehicle positions are obscured at each timestep with probability \(p \in [0,1]\). These agents with partial observability can be seen as self-driving agents operating in environments with extreme conditions that limit their vision of the road and of the other cars, similar to the scenarios described in the introduction, or to self-driving vehicles with noisy sensors. By introducing these controlled variations in observability, IGDrivSim provides a structured benchmark for studying the impact of partial observability on imitation learning in autonomous driving.  

\subsection{Metrics}
\label{subsec:metrics}

To evaluate the imitator's driving policy, we use commonly used safety metrics—\textit{overlap} (collision detection) and \textit{off-road} (leaving the road)—as well as \textit{log divergence} to measure deviation from the expert. We extend these per-step metrics to the full scenario with \textit{overlap rate} (any collision), \textit{off-road rate} (any off-road event), and \textit{maximal log divergence} over the trajectory. 

\subsection{Reward}
To compute the RL component of the combined loss defined in Equation~\ref{eq:loss}, we utilize the reward function defined in the Waymax simulator, which is the sum of safety penalties incurred during each time step, specifically due to off-road driving and vehicle overlap: $r_t = -\mathds{1}_\text{overlap}(t) - \mathds{1}_\text{off-road}(t),$
where $\mathds{1}_\text{overlap}(t)$ is an indicator function that equals one if the ego vehicle collides with another vehicle at timestep $t$, and $\mathds{1}_\text{off-road}(t)$ equals one if the ego vehicle is driving off the road at timestep $t$.

\section{Experiments}


Our experiments highlight BC’s limitations in training driving policies in settings with imitation gaps, compared to a method that combines BC and RL. Each experiment in \Cref{subsec:results} ran on an 8-GPU (NVIDIA GTX 1080) cluster for $\approx$ 2 days, using 230 GB RAM.

\subsection{Training}
\label{subsec:training}

\noindent \paragraph{\textbf{Architecture}} For BC training, we use an MLP to encode the object features and a polyline encoder from \cite{shi2023motion} for the roadgraph. After applying layer norms, the embeddings are concatenated and passed to an RNN with a final softmax to predict action probabilities. Following \cite{gulino2023waymax}, the action space is discretized into 256 bins. The RNN enables handling the partial observability in the environment by maintaining and updating a hidden state that captures information about the environment over time despite limited observations at each timestep. For RL, we used the same architecture but added an MLP value head after the RNN~cell.

\begin{table*}[t]
    \centering
    \begin{minipage}{0.58\textwidth} 
        \caption{\small \textbf{Regularizing BC with RL improves safety metrics.} Mean trajectory and per-step metrics for BC and BC-RL policies under varying partial observability in IGDrivSim. Blue cells denote BC-only policies, orange cells BC-RL ($w_{BC}=1, w_{RL}=0.05$). The mean is averaged over three seeds and significant differences are in bold.}
        \centering
        \label{tab:results}
        \resizebox{\textwidth}{!}{%
        \begin{tabular}{lccccccc|c}
\toprule
    \textbf{Metric} & \textbf{Full obs} & \textbf{Gaussian} & \textbf{Circular} & \textbf{Circular} & \textbf{Conic FoV} & \textbf{Random}\\
    & & \textbf{noise ($\sigma$=3)} & \textbf{FoV (r=4)} & \textbf{FoV (r=6)} & \textbf{(r=4, $\alpha=\frac{2}{3}\pi$)} & \textbf{masking (p=0.7)}\\
\midrule
log & \cellcolor{lightgreen}\textbf{3.67} & \cellcolor{lightgreen}\textbf{3.69} & \cellcolor{lightgreen}4.18 & \cellcolor{lightgreen}\textbf{3.87} & \cellcolor{lightgreen}4.60 & \cellcolor{lightgreen}\textbf{3.68} \\
divergence& \cellcolor{lightblue}3.79 & \cellcolor{lightblue}3.85 & \cellcolor{lightblue}4.12 & \cellcolor{lightblue}3.94 & \cellcolor{lightblue}\textbf{4.19} & \cellcolor{lightblue}3.81\\
\midrule
max log & \cellcolor{lightgreen}\textbf{10.56} & \cellcolor{lightgreen}\textbf{10.72} & \cellcolor{lightgreen}12.54 & \cellcolor{lightgreen}\textbf{11.51} & \cellcolor{lightgreen}14.08 & \cellcolor{lightgreen}\textbf{10.68} \\
divergence & \cellcolor{lightblue}11.00 & \cellcolor{lightblue}11.23 & \cellcolor{lightblue}12.28 & \cellcolor{lightblue}11.74 & \cellcolor{lightblue}\textbf{12.39} & \cellcolor{lightblue}11.16\\
\midrule
overlap & \cellcolor{lightgreen}\textbf{4.80} & \cellcolor{lightgreen}\textbf{4.66} & \cellcolor{lightgreen}\textbf{5.28} & \cellcolor{lightgreen}\textbf{5.09} & \cellcolor{lightgreen}\textbf{3.70} & \cellcolor{lightgreen}\textbf{4.95} \\
(\%) & \cellcolor{lightblue}5.06 & \cellcolor{lightblue}5.02 & \cellcolor{lightblue}5.45 & \cellcolor{lightblue}5.31 & \cellcolor{lightblue}5.73 & \cellcolor{lightblue}5.30 \\
\midrule
overlap & \cellcolor{lightgreen}\textbf{21.17} & \cellcolor{lightgreen}\textbf{20.94} & \cellcolor{lightgreen}24.25 & \cellcolor{lightgreen}\textbf{22.73} &
\cellcolor{lightgreen}24.65 & \cellcolor{lightgreen}\textbf{21.70}\\
rate (\%) & \cellcolor{lightblue}22.24 & \cellcolor{lightblue}22.26 & \cellcolor{lightblue}24.30 & \cellcolor{lightblue}23.47 & \cellcolor{lightblue}25.16 & \cellcolor{lightblue}23.00 \\
\midrule
offroad & \cellcolor{lightgreen}\textbf{2.13 }& \cellcolor{lightgreen}\textbf{2.39} & \cellcolor{lightgreen}\textbf{0.77} & \cellcolor{lightgreen}\textbf{1.13} &\cellcolor{lightgreen}\textbf{1.10} & \cellcolor{lightgreen}\textbf{2.20} \\
(\%) & \cellcolor{lightblue}2.54 & \cellcolor{lightblue}2.96 & \cellcolor{lightblue}2.15 & \cellcolor{lightblue}1.68 & \cellcolor{lightblue}2.39 & \cellcolor{lightblue}2.71 \\
\midrule
offroad & \cellcolor{lightgreen}6.69 & \cellcolor{lightgreen}\textbf{7.25} & \cellcolor{lightgreen}\textbf{4.28} & \cellcolor{lightgreen}\textbf{4.49} & \cellcolor{lightgreen}7.00 & \cellcolor{lightgreen}\textbf{6.85}\\
rate (\%) & \cellcolor{lightblue}7.78 & \cellcolor{lightblue}8.73 & \cellcolor{lightblue}7.18 & \cellcolor{lightblue}5.73  & \cellcolor{lightblue}8.34 & \cellcolor{lightblue}7.96 \\
\bottomrule
        \end{tabular}%
        }
    \end{minipage}
    \hfill
\begin{minipage}{0.41\textwidth} 
    \centering
    \vspace{-3mm}
    \includegraphics[width=\textwidth]{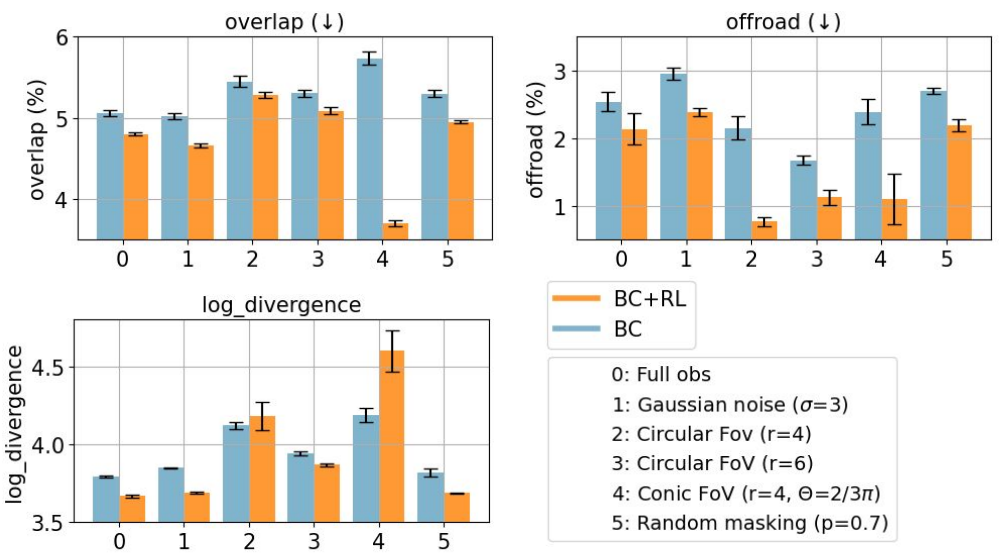}
    \captionof{figure}{\small Comparison of per-step metrics—log divergence, off-road, and overlap—between policies trained with BC (blue) and the combined BC-RL (orange) using parameters ($w_{BC}=1, w_{RL}=0.05$). The metrics are averaged over three seeds, with 95\% confidence intervals shown as error bars, for imitators under different partial observability.}
    \label{fig:results_waymax}
\end{minipage}
\vspace{-4mm}
\end{table*}

\noindent \paragraph{\textbf{Policy Update}} During training with the BC-RL combined loss, each scenario includes a logged trajectory—where the environment is updated by expert actions—and a simulated trajectory—where it is updated by imitator actions. The BC loss component is computed on the logged trajectory using the equation defined in Section~\ref{subsec:preliminaries}. To reduce distribution shift, similarly to \cite{seff2023motionlm}, we discretize the expert's demonstrations using the expert agent introduced in Waymax. The RL loss component (PPO loss), is computed on the simulated trajectory where reward has been recorded at each timestep. Finally, the model is updated using the Adam optimizer with a cosine decay learning rate, ranging from $10^{-3}$ to $0$, applied to the combined loss defined in Equation~\ref{eq:loss}.

\noindent \paragraph{\textbf{Training Set}} We filter the WOMD training dataset, which initially contained 487,002 driving scenarios, to retain only those where the ego car demonstrates interactive behavior. A scenario is classified as interactive if the ego car's trajectory intersects with another object's trajectory during the scenario. This reduces the training dataset by approximately $50\%$. To further reduce the dataset, we exclude scenarios where the ego car's mean speed throughout the trajectory is less than 0.7~m/s. The final dataset consists of 152,808 scenarios, representing approximately $31\%$ of the original WOMD training set.

\noindent \paragraph{\textbf{Test Set}} As the logged trajectories of the WOMD test set have not been released, we evaluate the performances of our trained policies on the validation set, distinct from the training set, which is composed of 44,097 driving scenarios. 

\subsection{Results}
\label{subsec:results}

To show that BC alone fails to learn efficient and safe behavior under imitation gaps—widened in IGDrivSim by observability constraints—we train imitators with varying observability levels. We then compare the policies learned through BC on human expert demonstrations with those trained using a combined BC and RL loss. As our focus is on IL of a single agent; while Waymax allows for controlling all the agents of the scene, we restrict the control to the ego car from which the scene was recorded. 

In the toy environment, we access a reward function that fully defines the desired behavior, allowing us to approximate the Bayes-optimal policy and thus characterize BC failure as the performance gap between the BC policy and the approximate Bayes-optimal policy under the imitator's POMDP. However, in the Waymax environment, such a reward function is inaccessible. Instead, by using the BC-RL combined loss, we show that there exists a superior policy under the imitator's POMDP that BC alone fails to recover. This approach establishes a lower bound on the performance gap between the BC policy and the Bayes-optimal policy in the presence of the imitation gap.



We evaluate the imitator under full scene observability and five different limited observability constraints: (1) limited field of view—either circular regions with radii of 6 and 4 meters, or a conical region with an angle of \( \theta = \frac{2}{3} \pi \) and a radius of 4 meters; and (2) noisy perception—where Gaussian noise (standard deviation: 3 meters) affects the imitator’s position, or the vehicle’s position is randomly masked with probability \( p = 0.7 \) (\cref{subsec:obs_waymax}). These agents' perceptions differ significantly from human expert perception, thereby introducing an imitation gap in the training process. Notably, even the agent with full observability relies solely on sensor-recorded data, which inherently differs from true human perception that could include additional factors like traffic sounds.

\Cref{fig:results_waymax} compares per-step metrics (log divergence, overlap, and off-road) for policies trained with BC and BC-RL across the six imitators, while \Cref{tab:results} provides the detailed values along with the trajectory metrics. For all the imitators, balancing the BC loss with the RL loss using the simple penalty reward helps the agent learn a significantly safer policy, reducing both overlap and off-road incidents. This demonstrates that using the BC-RL combined loss allows for better behavior learning under the imitator's POMDP than BC alone. Although \cite{lu2023imitation} highlights similar findings, they do not address the imitation gap, which we consider a key factor in explaining these results. Moreover, we have open-sourced our code, making these findings more accessible to the research community. 



Furthermore, as shown in Figure~\ref{fig:results_waymax}, the policies trained with the combined BC-RL loss not only improve safety metrics but also achieve lower log divergence—indicating closer alignment with the expert’s policy—compared to those trained solely with BC. The only exception is the imitator with a conic field of view (\(\theta = \frac{2}{3} \pi\), radius = 4 meters), which has the strictest observability constraint in our experiment. This indicates that under this partial observability, the policy trained with the combined loss diverges from the expert’s but performs better (i.e., more safely) within its own POMDP. RL regularization allows the imitator to learn behaviors not demonstrated by the human expert but effective in reducing overlap and off-road incidents based on its perception of the cars and road environment, demonstrating how RL helps narrow the imitation gap.

Qualitatively, the agent trained with the BC-RL combined loss shows greater reactivity to changes in its receptive field. For example, it engages in heavy braking when the road lines of other vehicles enter its view. In contrast, in several manually inspected scenarios, expert human drivers, having broader perception, anticipate road limits and nearby vehicles earlier, leading to smoother braking and overall fluid driving behavior. The RL regularization of the BC loss enables the agent to learn effective behaviors within the imitator's POMDP that were not observed in the expert demonstrations. 

\section{Conclusion}
\label{sec:conclusion}

Our work highlights the challenges of training self-driving agents from human demonstrations when an imitation gap exists caused by differences in perception between human experts and self-driving car imitators. We introduce IGDrivSim, a benchmark designed to study how this gap affects IL from human expert demonstrations. Our results show that self-driving cars trained only with IL often struggle to learn safe and efficient driving behaviors under their own perception of the environment when there is a significant imitation gap. However, we also found that combining IL with RL--using simple penalties for dangerous driving behaviors--can help mitigate these issues and improve overall performance. We release open-source code and motion prediction baselines trained on the Waymax simulator, for which no previous open-source baselines were available. We hope that IGDrivSim will enhance understanding of IL from human driving demonstrations and lead to the development of safer and more effective self-driving policies tailored to the specific sensors of self-driving cars.

\bibliographystyle{IEEEtran}
\bibliography{references}

\end{document}